\documentclass{article}
\usepackage[preprint,nonatbib]{neurips_2018}

\usepackage{longtable}

\usepackage{booktabs}

\usepackage{mathrsfs}
\usepackage{algorithm}
\usepackage{algorithmic}
\usepackage{wrapfig}
\usepackage{xcolor}

\usepackage{multirow}
\usepackage{lscape}
\usepackage{tabulary}
\usepackage{bbm}

\usepackage[utf8]{inputenc} 
\usepackage[T1]{fontenc}    
\usepackage{hyperref}       
\usepackage{url}            
\usepackage{booktabs}       
\usepackage{amsfonts}       
\usepackage{nicefrac}       
\usepackage{microtype}      

\usepackage{amsmath}
\usepackage{algorithm, algorithmic}
\usepackage{amsfonts, amssymb}
\usepackage{multirow}
\usepackage{mathtools}
\usepackage{arydshln}
\usepackage{refcount}
\usepackage{listings}
\usepackage{hyperref}

\newtheorem{theorem}{Theorem}[section]

\newtheorem{assumption}[theorem]{Assumption}
\newtheorem{proof}[theorem]{Proof}

\usepackage{times}
\usepackage{soul}
\usepackage{url}

\title{Dual Behavior Regularized Reinforcement Learning}

\author{%
  Chapman Siu \\
  Faculty of Engineering and Information Technology \\
  University of Technology Sydney, Australia \\
  \texttt{chpmn.siu@gmail.com} \\
  \And
   Jason Traish \\
  Faculty of Engineering and Information Technology \\
  University of Technology Sydney, Australia \\
  \texttt{Jason.Traish@uts.edu.au} \\
  \AND
  Richard Yi Da Xu \\
  Faculty of Engineering and Information Technology \\
  University of Technology Sydney, Australia \\
  \texttt{YiDa.Xu@uts.edu.au} \\
}

\begin{document}

\maketitle

\begin{abstract}
Reinforcement learning has been shown to perform a range of complex tasks through interaction with an environment or collected leveraging experience. However, many of these approaches presume optimal or near optimal experiences or the presence of a consistent environment. In this work we propose dual, advantage-based behavior policy based on counterfactual regret minimization. We demonstrate the flexibility of this approach and how it can be adapted to online contexts where the environment is available to collect experiences and a variety of other contexts. We demonstrate this new algorithm can outperform several strong baseline models in different contexts based on a range of continuous environments. Additional ablations provide insights into how our dual behavior regularized reinforcement learning approach is designed compared with other plausible modifications and demonstrates its ability to generalize. 
\end{abstract}

\section{Introduction}

Reinforcement learning typically focuses on accumulating the maximum cumulative discounted reward as part the environment. In the typical reinforcement learning scenario where agents have access to their environment, active data collection is encouraged, and in doing so, algorithms are developed which discard or ignore sub-optimal behavior of an agent \cite{sil,abdolmaleki2018maximum}. Instead, a rather greedy approach is typically used in reinforcement learning, which focuses on mode-seeking behavior, and over modeling the density of the whole space \cite{modeseeking,vaepolsearch}. To this end, approaches generally have a bias towards high rewards, whilst penalizing and even outright `forgetting' about bad experiences or rewards. This is often formulated as {\em optimism} in reinforcement learning, which is typically formulated in the regret settings \cite{arm}. 

Concurrently, there has been much interest in offline reinforcement learning, whereby the reinforcement learning approach is restricted to only a static offline dataset of experiences \cite{brac,bear,awac}. This setting is of particular interest when it is difficult to test or deploy a policy, or where there are practical concerns in updating the policy in an online manner \cite{litoward}. For example, recommendation systems \cite{Bennett07thenetflix}, autonomous driving \cite{yu2020bdd100k} and health applications \cite{murphy2001marginal}, are arenas which contain plentiful amounts of capture data, however deploying new policies may not be done in a live or active manner, rather only after extensive testing and evaluation. 

Due to the ``greedy'' nature of many reinforcement learning algorithms, they typically perform poorly in the offline setting. Off-policy reinforcement learning algorithms such as  Deep Q-Networks (DQN) \cite{mnih2013playing}, Deep Deterministic Policy Gradient (DDPG) \cite{lillicrap2015continuous}, and Soft Actor Critic (SAC) \cite{sac} have been shown to be sensitive to the distribution of experience in the replay buffer \cite{fudiag}. 

We are interested in developing methods which address both of these challenges, and understanding conditions where a general reinforcement learning framework could be used across a variety of reinforcement learning problems, whether it be in an adversarial, offline, or partially observable setting. That is, can we develop methods which can be trained offline, and online but still remain competitive to approaches which generally approach one or the other?

Our contribution is a new model-free deep reinforcement learning algorithm based on behavior regularized actor critics that also estimates an advantage-like function, approximating a quantity called the counterfactual regret. This approach combines ideals from several algorithms, namely Behavior Regularized Actor Critic (BRAC) \cite{brac}, and is a policy variation of the Advantage-based Regret Minimization (ARM) \cite{arm}. 

We evaluate our approach on a range of continuous control environments: the standard PyBullet environments and the PyBullet offline reinforcement learning environment \cite{d4rl}. In our experiments, we find that our method is more robust and generalizes better over prior methods across different reinforcement learning contexts.

\section{Related Work}


One challenge in the merging of offline reinforcement learning and the off-policy learning is that it can be confirmed empirically that in the offline setting, policy approaches such as TD3 and DDPG fail to learn a good policy, even when the policy is learned by a single behavior with or without noise added to the behavior policy \cite{brac}. This arises by erroneous generalization of the state-action value function (Q-value function) learned with function approximators. This has been addressed by Behavior regularized Actor Critic approaches \cite{bear,brac}, which aims to regularize the learned policy towards the behavior policy based on the intuition that unseen state-action pairs are more likely to receive overestimated Q-values. These approaches may used different metrics to regularize the behavior policy with the target policy, such as the usage of Maximum Mean Discrepancy (MMD) \cite{bear} or through Kullback-Leibler (KL) divergence \cite{brac}. 

Concurrently, other approaches which aim to regularize policies \textit{implicitly} have been studied. The advantage weighted actor-critic (AWAC) algorithm, trains an off-policy critic and an actor with an implicit policy constraint without the use of a behavior policy in the offline reinforcement learning setting. This is achieved through learning a policy that maximizes the value of the critic, but implicitly restricting the policy distribution to stay close to the data observed during the actor update \cite{awac}. Another approach is the Munchhausen Reinforcement Learning (MRL) \cite{mrl}, which is a more generalized approach for any Q-learning algorithm that aims to augment the immediate reward in the Q-value target update step with the scaled log-policy. That is the augmented reward in the form $r^{mrl}_{t} := r_{t} + \tau \log \pi(a | s)$ is used, where $\tau \in [0, 1]$ is the scaling factor. This augmentation has been shown to implicitly perform KL regularization between successive policies in the policy update. 

In a related setting, deep reinforcement learning algorithm based on counterfactual regret minimization using advantage explores other ways to yield robust algorithms towards the update rule for a modified cumulative Q-function \cite{arm}. This form of advantage function updating resembles our approach towards the dual behavior aspect of our actor critic algorithm. A significant difference is that our approach can be used in continuous action space environments. By extension this approach also results in a multi-step update reinforcement learning algorithm shown in Self-Imitation Learning (SIL) \cite{sil}, which explicitly updates using the advantage in a separate step, rather than completely inline.  

\section{Dual Behavior Regularized Reinforcement Learning}

In this section, we provide some background on Behavior Regularized Actor critic, describe Dual Behavior Regularized Reinforcement Learning (DBR) in detail and give some intuition for why DBR works. 

\subsection{Markov Decision Process}

We represent the environment as a Markov decision process (MDP) defined by a tuple $(\mathcal{S}, \mathcal{A}, P, R, \rho_0, \gamma)$, where $\mathcal{S}$ is the state space, $\mathcal{A}$ is the action space, $P(s^\prime \vert s, a)$ is the transition distribution, $\rho_0$ is the initial state distribution, $R(s, a)$ is the reward function, and $\gamma \in (0, 1)$ is the discount factor. For convenience, we define the replay buffer to be the tuple $\mathcal{D} = (s, a, r, s^\prime)$. The goal in reinforcement learning is to find a policy $\pi(a \vert s)$ that maximizes the expected cumulative discounted rewards. 



\subsection{BRAC and BEAR}

First, we consider behavior regularized approaches to the \textit{offline} reinforcement learning problem, which is described by BRAC \cite{brac} and BEAR \cite{bear}. Consider the Soft Actor Critic (SAC) framework \cite{sac} where the Value function of the policy is frames as 

\begin{equation}
V^\pi(s) = \sum_{t=0}^\infty \gamma^t \mathbb{E}_{s_t \sim P_t^\pi(s)}[R^\pi (s_t) + \alpha H(\pi(\cdot | s_t))]
\end{equation}
This is extended in the BRAC \cite{brac} and BEAR \cite{bear} framework 

\begin{equation}
V^\pi(s) = \sum_{t=0}^\infty \gamma^t \mathbb{E}_{s_t \sim P_t^\pi(s)}[R^\pi (s_t) - \alpha D(\pi(\cdot | s_t), \beta(\cdot | s_t))]
\end{equation}

Where $D$ is the choice of divergence function and $\beta$ represents the behavior policy used to generate the static dataset of transitions $\mathcal{D}$, and as such $\beta$ is always well-defined even if the dataset was collected by multiple, distinct behavior policies. As such we do not assume direct access to $\beta$ and it is commonly approximated as $\hat{\beta} := \mathop{\mathrm{argmax}}_{\hat{\pi}} \mathbb{E}_{(s, a, r, s^\prime) \sim \mathcal{D}}[\log \hat{\pi}(a, \vert s)]$. In the BRAC framework it is Kullback-Leiber (KL) divergence and in the BEAR framework it is Maximum Mean Discrepancy (MMD). 
The BEAR framework adds an additional threshold $\epsilon \geq 0$ to the kernel MMD distance, i.e. $\mathbb{E}_{s \sim \mathcal{D}}[D(\pi(\cdot | s_t), \beta(\cdot | s_t))] < \epsilon$ which allows the BEAR algorithm to control the level of constraint the policy has relative to the behavior policy, which in turn allows the BEAR algorithm to tune faster or slower when performing fine-tuning.

\subsection{ARM and SIL}

Next, we consider two approaches which use variations of the clipped advantage in the formulation of the agent policy. Self-Imitation Learning (SIL) \cite{sil} builds on top of SAC, whereby at every time step, an additional update is performed using the clipped advantage to update the actor-critic model. The clipped advantage is in the form $ (R^{sil}-V_\theta(s))_{+}$, where $R^{sil} = \sum_{k=0}^\infty \gamma^k r_k$ is the discounted sum of rewards, where $r_k$ and $\gamma$ is the reward at time $k$ and the discount rate respectively, $(\cdot)_{+} = \text{max}(\cdot, 0)$ and $V_\theta$ is the value function of an actor-critic policy parameterized by $\theta$. In comparison, the clipped advantage used in ARM is in the form $(Q_{\theta}(s, a)-V_\theta(s))_{+}$, with the cumulative clipped advantage at training iteration $T$ being $A_T^{+} = (A_{T-1}^{+})_{+} + (Q_{\theta}(s, a)-V_\theta(s))$, where the key difference is ARM uses the $Q$ function instead of the cumulative discounted reward. ARM does not require a separate update step in training the policy, but instead uses the clipped advantage directly to train the policy. However if this approach was extended to policy gradient approaches, as the formulation of the policy update can yield $\log(0)$, ARM models are restricted to Q-learning approaches in the discrete action space \cite{arm}. This is not a restriction for SIL approach, as the any update which would have yielded a $\log(0)$ is simply discarded and not used in the additional update step that is performed.

\subsection{From ARM and BEAR to DBR}
\label{c4_sec}

\footnotetext{Image taken from ``gameplay of Connect-4'', source: Wikipedia -  \url{https://en.wikipedia.org/wiki/Connect_Four}}

In constructing Dual Behavior Regularized Reinforcement Learning (DBR), we address some limitations of BEAR and ARM. As part of the BEAR formulation the level of constraint is a fixed constant $\epsilon$ which is set to allow the rate of which the BEAR policy can be updated during fine tuning. For ARM, the update formulation is restricted to Q-learning and discrete action spaces. To address these deficiencies we use split the approximation of the behavioral policy into two components, being the ``clipped advantage'', $(R-V_\theta(s))_{+}$ and the ``{\em negative} clipped advantage'', $(R-V_\theta(s))_{-}$, where the functions are defined as $(\cdot)_{+} = \text{max}(\cdot, 0)$ and $(\cdot)_{-} = \text{min}(\cdot, 0)$. We define $\hat{\beta}_{+} := \mathop{\mathrm{argmax}}_{\hat{\pi}} \mathbb{E}_{(s, a, r, s^\prime) \sim \mathcal{D}_{+}}[\log \hat{\pi}(a, \vert s)]$ and $\hat{\beta}_{-} := \mathop{\mathrm{argmax}}_{\hat{\pi}} \mathbb{E}_{(s, a, r, s^\prime) \sim \mathcal{D}_{-}}[\log \hat{\pi}(a, \vert s)]$, where $D_{+}, D_{-}$ are subsets of the replay buffer where $R \geq V_\theta(s)$ and $R \leq V_\theta(s)$ respectively. This changes the formulation in the BRAC framework to be 

\begin{equation}
V^\pi(s) = \sum_{t=0}^\infty \gamma^t \mathbb{E}_{s_t \sim P_t^\pi(s)}[R^\pi (s_t) - \alpha D(\pi(\cdot | s_t), \hat{\beta}_{+}(\cdot | s_t))]
\end{equation}

Whereby we restrict the policy improvement step to be dynamic, based on the behavior policy trained on the negative clipped advantage $\hat{\beta}_{-}$, which creates the dynamic constraint: $\mathbb{E}_{s \sim \mathcal{D}}[D(\pi(\cdot | s_t), \hat{\beta}_{+}(\cdot | s_t))] < \max(\epsilon, \mathbb{E}_{s \sim \mathcal{D}}[D(\pi(\cdot | s_t), \hat{\beta}_{-}(\cdot | s_t))])$. The intuition behind this update in a fine-tuning example, is that as the replay buffer updates with new experiences, we want the divergence between the behavior policies based experiences which yield higher advantage scores to be closer to our trained policy, than the corresponding behavior policy based on experiences which yield lower advantage scores. In this formulation, we avoid the $\log(0)$ issue present in ARM without discarding experiences as in SIL through training via the behavioral policies $\hat{\beta}_{+}, \hat{\beta}_{-}$, allowing all experiences to be used in a counterfactural regret minimization setting by optimizing over a surrogate objective $\hat{\beta}_{+}$, rather than the clipped advantage directly, that is we want to ensure the update is in the form $\mathop{\mathrm{argmin}}_{\pi} \mathbb{E}_{s \sim \mathcal{D}}[D(\pi(\cdot | s_t), \beta_{+}(\cdot | s_t)]$ \cite{arm,sac}. This allows the clipped advantage to be approximated through the behavioral policy $\beta_{+}$, that is $\beta_{+} \approx A_{+}$

As such our approach is essentially ARM with different design choices around the surrogate objective in order to facilitate ARM-like policy gradient, which was deficient in the original ARM algorithm \cite{arm}. 

These changes to the BEAR algorithm still allow DBR to maintain a concentrability coefficient bound as described in BEAR based on the marginal state distribution under the dual data distribution  $\mathcal{D}_{+}$ and $\mathcal{D}_{-}$. We are interested in sets of policies $\Pi$, constrained by support sets where $\Pi_\epsilon = \{ \pi | \pi(a | s) = 0 \text{ whenever } \beta_{+}(a|s) < \max(\epsilon, \beta_{-}(a|s)\}$, (i.e. $\Pi$ is the set of policies that have support in the probable regions of the behavior policy based on the clipped advantage, but away from the behavior policy based on the negative clipped advantage). Defining $\Pi_\epsilon$ in this way allows us to bound the concentrability coefficient.

\begin{assumption}
{\em(Concentrability: \cite{bear})}. Let $\rho_0$ denote the initial state distribution and $\mu(s, a)$ denote hte distribution of the training data over $\mathcal{S} \times \mathcal{A}$, with marginal $\mu(s)$ over $\mathcal{S}$. Support these exist coefficients $c(k)$ such that for any $\pi_1, \dots, \pi_k \in \Pi$ and $s \in \mathcal{S}$:

\begin{equation}
\rho_0 P^\pi_1 P^\pi_2\dots P^\pi_k (s) \leq c(k) \mu(s)
\end{equation}

where $P^\pi_i$ is the transition operator on states induces by $\pi_i$. Then, define the concentrability coefficient $C(\Pi)$ as 

\begin{equation}
C(\Pi) := (1- \gamma)^2 \sum_{k=1}^\infty k \gamma^{k-1}c(k)    
\end{equation}

\end{assumption}

\begin{theorem}
Let $\mu_{\beta_i}$ denote the the data distribution generated by behavior policy $\beta_i$. Let $\mu_{\beta_i}(s)$ be the marginal state distribution under the corresponding data distribution. Define $\Pi_\epsilon = \{ \pi | \pi(a | s) = 0 \text{ whenever } \beta_{+}(a|s) < \max(\epsilon, \beta_{-}(a|s)\}$ and let $\mu_{\Pi_\epsilon}$ be the highest discounted marginal state distribution starting from the initial state distribution $\rho$ and following policies $\pi \in \Pi_\epsilon$ at each time step thereafter. Then, there exists a concentrability coefficient $C(\Pi_\epsilon)$ which is bounded:

$$C(\Pi_\epsilon) \leq C(\beta_{+}) \cdot \left(1 + \frac{\gamma}{(1-\gamma)f(\epsilon)}(1-\epsilon)\right)$$

where $f(\epsilon) := \max_{\tilde{\beta} \in \{\beta_{+}, \beta_{-}\}}\min_{s \in S, \mu_{\Pi_\epsilon}(s) > 0}[\mu_{\tilde{\beta}}(s)] >0$.
\end{theorem}

\begin{proof}

This follows from the result presented in BEAR, observing that despite the addition of dual behavior policies $f(\epsilon)$ remains as a constant that depends on $\epsilon$. Similar to BEAR \cite{bear}, for notational clarity, we refer to $\Pi_\epsilon$ as $\Pi$ in this proof. The term $\mu_\Pi$ is the highest discounted marginal state distribution starting from the initial state distribution $\rho$ and following policies $\pi \in \Pi$. Formally, it is defined as:

$$\pi_\Pi := \max_{\{\pi_i\}_i:\forall i, \pi_i \in \Pi} (1-\gamma) \sum_{m=1}^\infty m \gamma^{m-1}\phi_0 P^\pi_1\cdots P^{\pi_m}$$

Now, we begin the proof of the theorem. We first note, from the definition of $\Pi, \forall s \in \mathcal{S}, \forall \pi \in \Pi, \pi(a|s) > 0 \Rightarrow \beta_{+}(a|s) > \epsilon$. It can be shown that the marginal state distributions of $\beta_{+}$ and $\Pi$, are bounded in total variation distance by $D_{TV}(\mu_{\beta_{+}} || \mu_\Pi) \leq \frac{\gamma}{1-\gamma}(1-\epsilon)$, where $\mu_\Pi$ is the marginal state distribution as defined above \cite{bear}. 

Furthermore, the definition of the set of policies $\Pi$ implies that $\forall s \in \mathcal{S}, \mu_\Pi(s) > 0 \Rightarrow \mu_{\beta_{+}} \geq f(\epsilon)$, 

where $f(\epsilon) > 0$ is a constant that depends on $\epsilon$ and captures the maximum of the minimum visitation probability of a state $s \in \mathcal{S}$ when rollouts are executed from the initial state distribution $\rho$ while executing behavior policy $\beta_{+}(a | s)$ or $\beta_{-}(a | s)$, under the constraint that only actions with $\beta_{+}(a|s) \geq \epsilon$ are selected fro execution in the environment. Notice that compared with BEAR, this constant $f(\epsilon)$ is greater than or equal to the minimum visitation probability over $\beta := \beta_{+} \cup \beta_{-}$. Combining it  with the total variation divergence bound, $\max_s \lVert \mu_{\beta_{+}}(s) - \mu_\Pi(s)\rVert \leq \frac{\gamma}{1-\gamma}(1-\epsilon)$, we get that

\begin{equation}
sup_{s\in S} \frac{\mu_\Pi(s)}{\mu_{\beta_{+}}}\leq 1+ \frac{\gamma}{(1-\gamma) f(\epsilon)}(1-\epsilon)
\end{equation}

We know that, $C(\Pi) := (1-\gamma)^2 \sum_{k=1}^{\infty} k \gamma^{k-1}c(k)$ is the ratio of the marginal state visitation distribution under the policy iterates when performing backups using the distribution-constrained operator and the data distribution $\mu = \mu_{\beta_{+}}$. Therefore as shown by \cite{bear},

\begin{equation}
\frac{C(\Pi_\epsilon)}{C(\beta_{+})} :=\sup_{s\in S} \frac{\mu_\Pi(s)}{\mu_{\beta_{+}}}\leq 1+ \frac{\gamma}{(1-\gamma) f(\epsilon)}(1-\epsilon)
\end{equation}
\end{proof} 

This bound provides theoretical guarantees, and proving simple and effective methods for implementing distribution constrained algorithms for dynamic approach to fine tuning. Furthermore, as we can choose actions in $\Pi_\epsilon$, the support of the training distribution and not restrict action selection to the policy distribution \cite{bear}, that is ensuring that the Q-value for each state is staying in the support of the training distribution. This provides a mechanism which can adaptively constrain the policies to the training distribution in an offline setting or for exploration in a structured way when fine tuning. 

\begin{figure*}
\centering
\begin{tabular}{c}
{\includegraphics[width=0.31\textwidth]{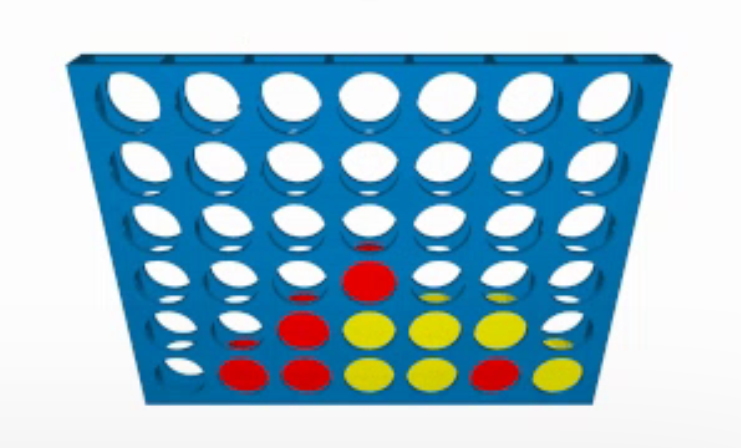} }

{\includegraphics[width=0.31\textwidth]{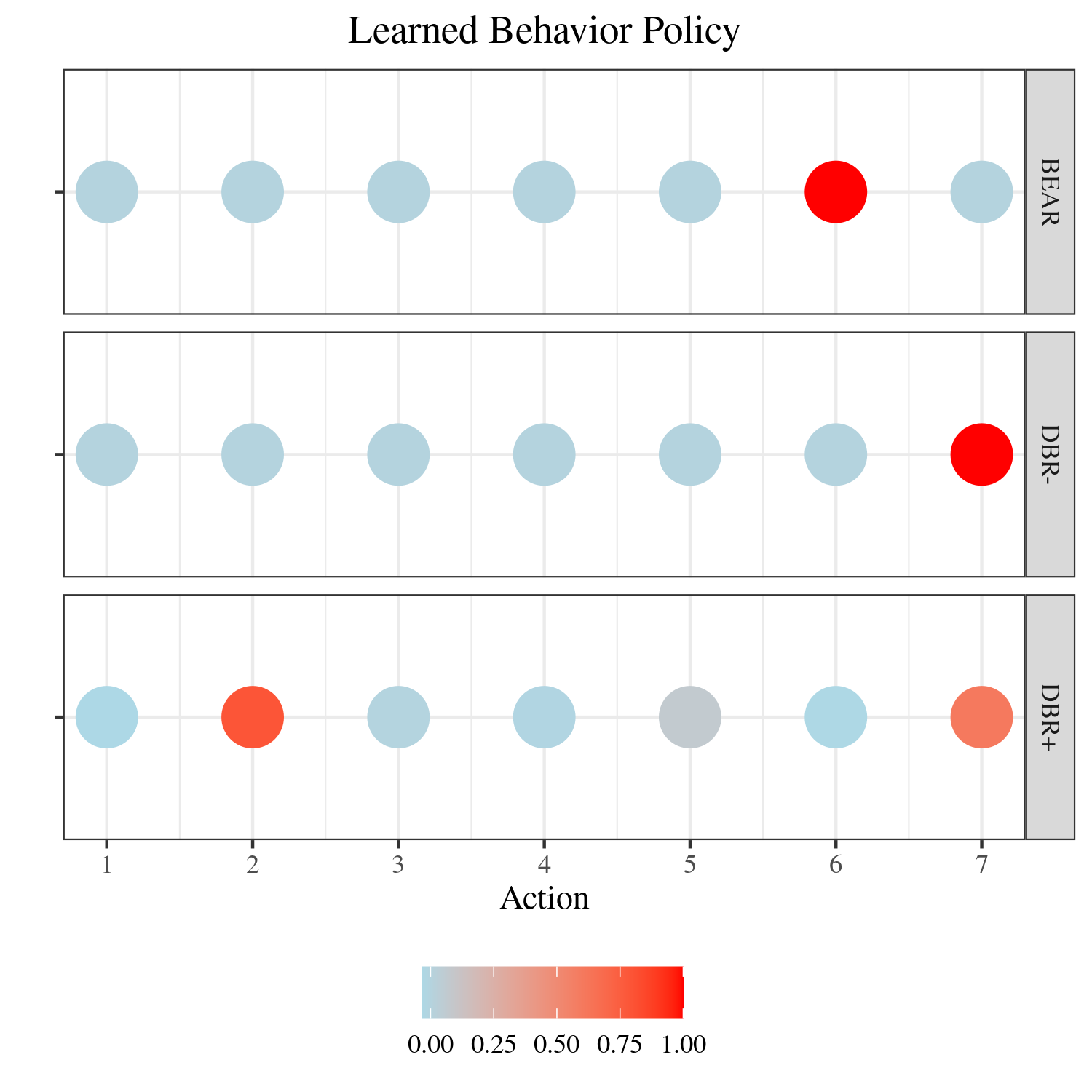} }
{\includegraphics[width=0.31\textwidth]{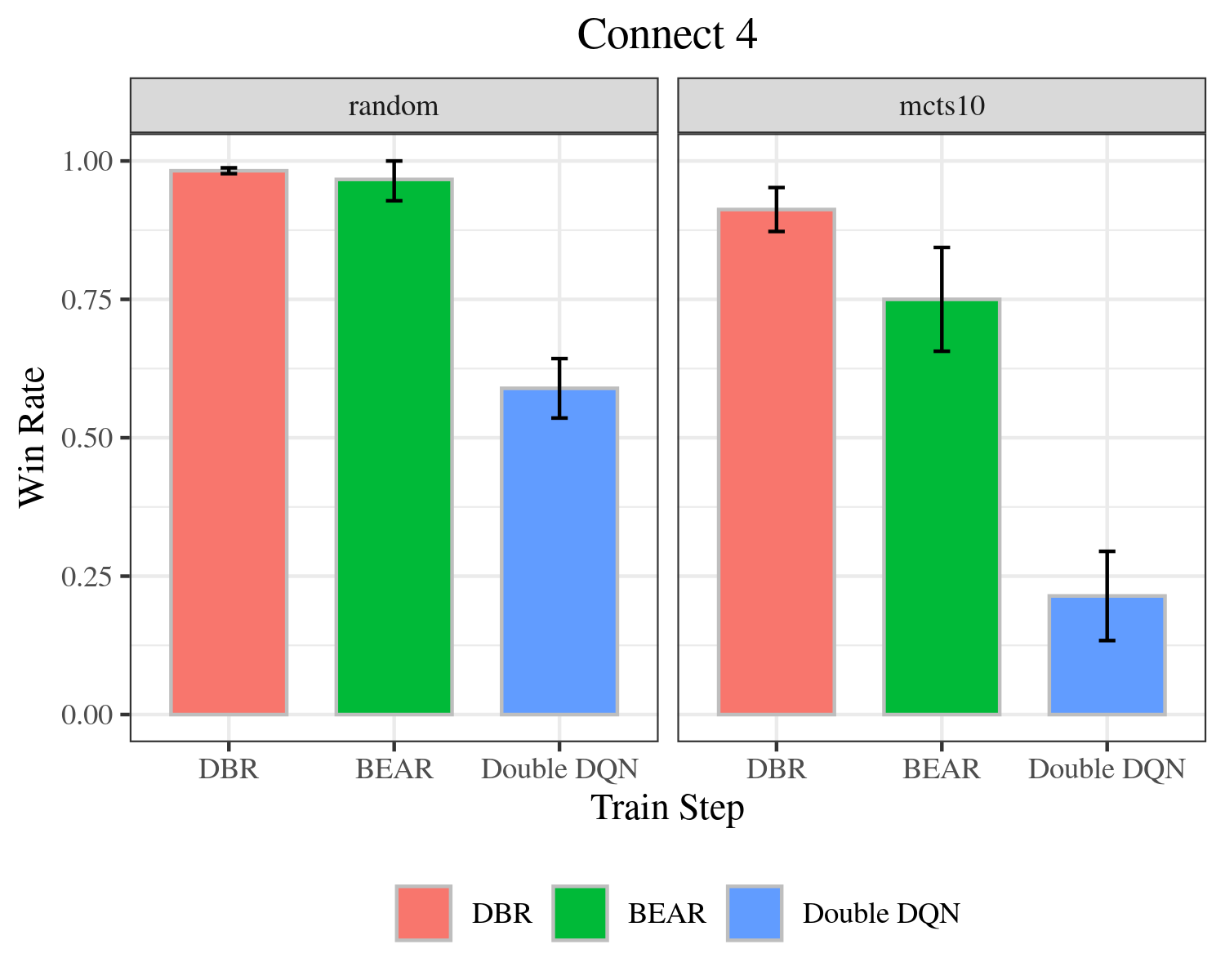} }
\end{tabular}
\caption{(Left) connect-4 Game with Red to Play. Optimal action is to play the right-most file block Yellow's winning move. (Center) Learned behavior policy (normalized) from BEAR, DBR (positive and negative) based on the board presented (Left). (Right) Evaluation performance against an (unseen) adversarial agent with MCTS depth 10, with Double DQN as the baseline approach. Error bars represent the standard deviation over these rollouts.}
\label{c4}

\end{figure*}

Compared with BEAR, this approach intentionally avoids of regions indicated by $\beta_{-}$ which is important particularly in environments with severe degenerative states. This is demonstrated in empirical experiences using Connect-4 game, shown in Fig. \ref{c4}, where it demonstrates DBR's ability to both learn key moves, and avoid key moves over BEAR and DQN.

\label{method:dbr}

It is natural to consider the scenarios where the dynamic constraint using $\beta_{-}$ is removed, and thereby using a fixed constraint as in BEAR: $\mathbb{E}_{s\sim \mathcal{D}}[D(\pi(\cdot | s_t), \hat{\beta_{+}}(\cdot | s_t))] < \epsilon$. This would form a variation of the BEAR algorithm where the behavior policy is trained using the ``clipped advantage'' like the ARM model \cite{arm}. Alternatively, if we train using the cumulative reward, rather than the Q value function, with a separate update step, we would have a SIL variation \cite{sil}. Both of these formulations are explored in the ablation experiments in Section \ref{ref:abl}. In constraining the actor policy, we also explore the Munchhausen Reinforcement Learning (MRL) approach, whereby we leverage the entropy regularization during the training of the policy as in SAC and MRL, however when we construct the Munchhausen reward, we instead use $r^{mrl}_{t} := r_t + \tau \log \beta_{+}(a | s)$, where $\tau \in [0, 1]$ is the scaling factor. These variations are explored in out ablation experiments in Section \ref{ref:abl}. 


\section{Experiments}

We hypothesise that DBR should perform well in both offline reinforcement learning and other reinforcement learning settings where it is forced to adapt. We conduct our experiments on the standard PyBullet environments and in the offline reinforcement learning, where the agents are trained on static dataset of experiences. Our evaluations use feedforward MLP. Our hyperparameters are based on the official implementations of their respective algorithms, with further information provided listed in the Appendix \ref{ref:hyper}. We are interested in comparing DBR with other offline methods, and online methods, primarily TD3, SAC, BEAR, AWAC. We also explore in the ablation section, variations of DBR which leverage self-imitation and Munchhausen extensions. We report the average evaluation return which were evaluated on 1000 evaluation episodes of the underlying unaltered environment, which is used to generate the average score and variance for the plots.

For all experiments, we use a 3-layer neural network of the same size for both the actor and critic networks, with the appropriate modifications according to each algorithm. The hyperparameters used in all experiments is shown below.

\begin{table}
\centering
\caption{Hyper-parameters used for RL experiments.}\label{app:tab:hyper}
\begin{tabular}{c|c}
\hline
\textbf{Hyperparameters}    & \textbf{Value} \\
\hline
Policy Hidden Sizes         & {[}256, 256, 256{]} \\
Policy Hidden Activation    & ReLU           \\
Target Network $\tau$          & 0.005          \\
Learning Rate               & 0.0003         \\
Batch Size                  & 256            \\
Replay Buffer Size          & 1000000        \\
Number of pretraining steps & 1000           \\
Steps per Iteration         & 1000           \\
Discount                    & 0.99           \\
Reward Scale                & 1              \\
\hline
\end{tabular}
\end{table}
\label{ref:hyper}

The policies used leverage the default hyper-parameters based on the official implementation of TD3, SAC and BEAR using their accompanying code \cite{bear}. For our algorithm DBR leveraged the same hyper-parameters as BEAR, including the usage of the $k=2$ for the $Q$-function ensembles for both BEAR and DBR as suggested in BRAC \cite{brac}. For the adversarial agents, the hyper-parameters used for TD3 is identical to the official implementation. In the ablation studies, the MRL, the suggested scaling term of $0.9$ is used. For AWAC implementation, we used the same hyper-parameters as SAC agents in order to make it comparable from an Actor-Critic model perspective, further details are provided in Table \ref{app:tab:hyper}

We evaluated on 1000 evaluation episode rollouts (separate from the train distributions) every training iteration and used the average score and variance for the plots and tables.

\subsection{Connect-4}

Connect-4 is a non-trivial board game of medium-high complexity, for which we constructed a heuristic agent using Monte-Carlo Tree Search \cite{mcts} as an agent to determine our approaches' ability to generalize beyond random actions and explore its representative power. We use Deep Reinforcement Learning with Double Q-learning (Double DQN) \cite{ddqn} as our benchmark approach due to the discrete nature of the environment. We modify both BEAR and DBR using discrete variation of SAC \cite{sac-discrete}. Figure \ref{c4} in Section \ref{c4_sec} demonstrates the ability for the various approaches to generalize, as we observe the level of degeneration when the agents move from the training environment to the evaluation environment. Our agents is trained and only have access against an opponent which makes random moves, and is evaluated on a separate unseen agent trained using Monte-Carlo Tree search with varying depths, all policies were trained for 2 million rollouts against a random agent. DBR $\beta_{-}$ successfully identifies the importance of the right most move whereby {\em not playing} will result in a loss, notice that leveraging $\beta_{+}$ alone in a greedy setting would still fail to identify this action, as it has a preference for a move on the left hand side, whereas BEAR unsuccessfully identifies the correct action.

\subsection{PyBullet Environments}


We train our agents on PyBullet environments {\em tabula rasa}. Although algorithms such as BEAR, AWAC are not necessarily designed to be used without the presence of offline data, these experiments demonstrate DBR's extensions to BEAR to enable dynamic control for exploration in this scenario, and explore the performance of AWAC where training is performed from the start; without the presence of offline datasets. This demonstrate DBR is superior to the offline reinforcement learning approaches (BEAR, AWAC), whilst comparable to baseline approaches (TD3, SAC). This demonstrates the reliance AWAC has in the implicit constrained on observed data, as it needs to build up the experience itself, which suggests its limitations in learning {\em tabula rasa}. In comparison, DBR demonstrates this compromise between the offline reinforcement learning techniques and the off-policy techniques, and able to balance between the two competing priorities. 

\begin{figure}
\centering
\begin{tabular}{c}
{\includegraphics[width=0.31\textwidth]{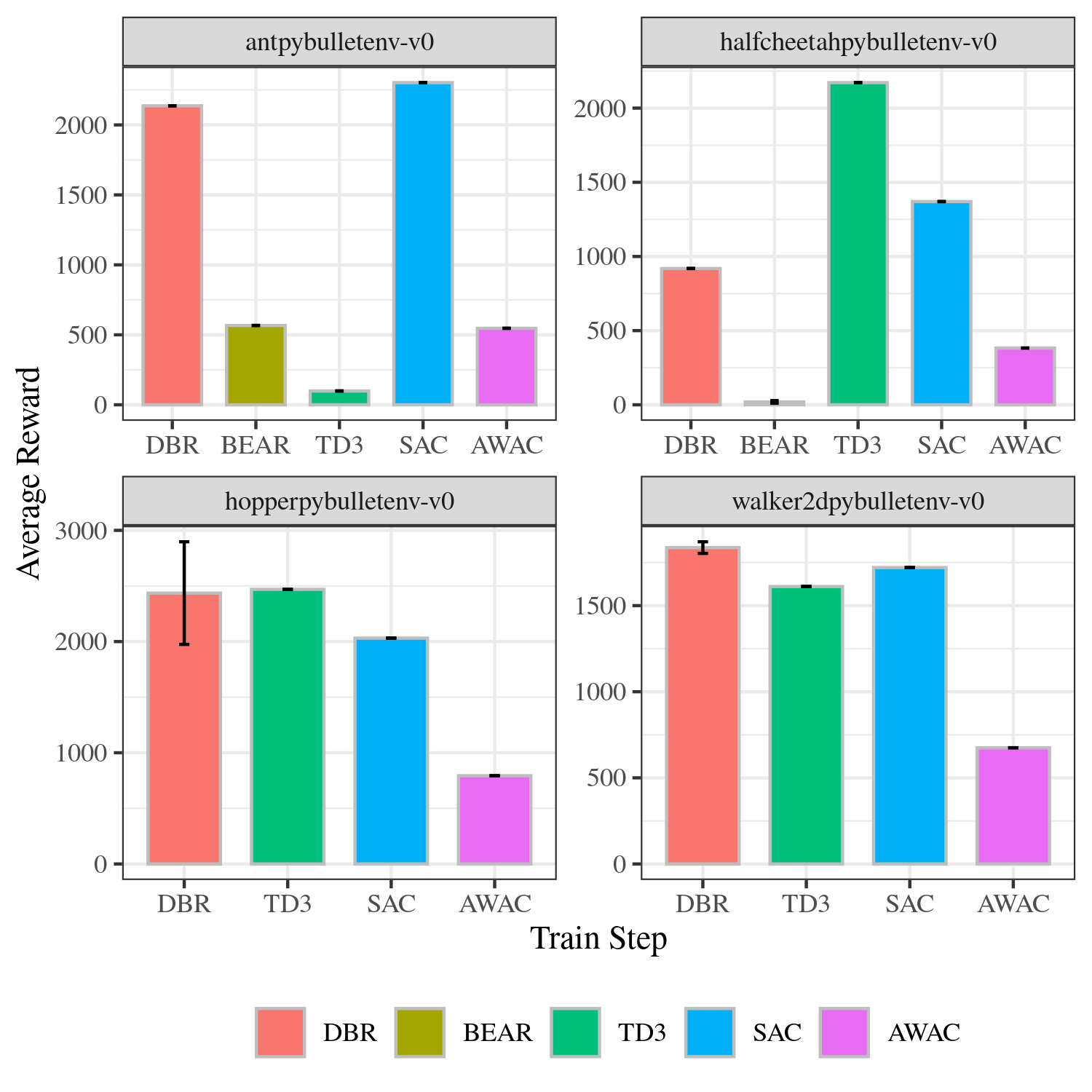}}
{\includegraphics[width=0.31\textwidth]{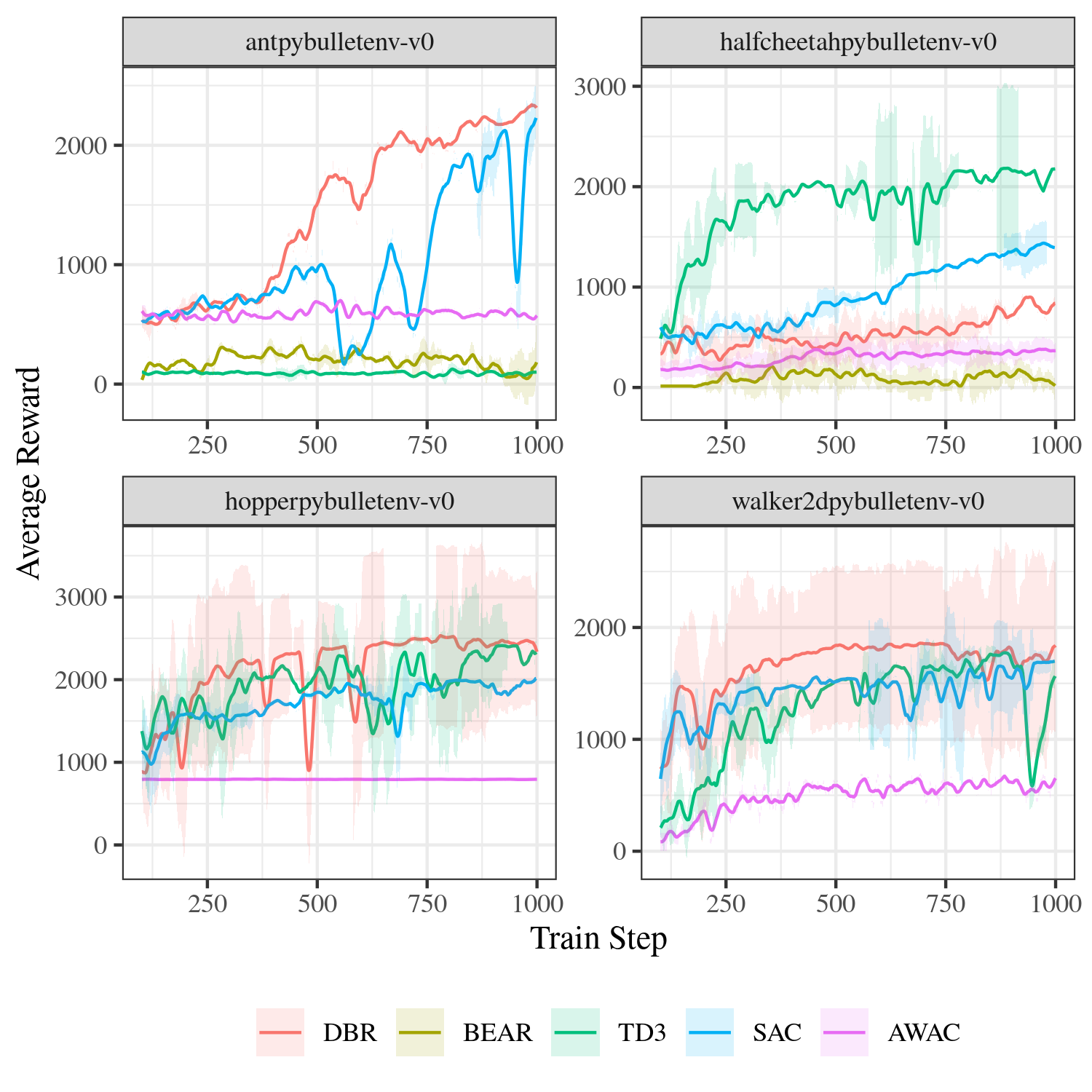}}

\end{tabular}
\caption{Performance over a range of PyBullet environments, when trained from scratch without the presence of offline dataset. We report the mean over the last 10 evaluation steps, where each evaluation point is average over 1000 episodes. Error bars represent the standard deviation over these rollouts.}
\label{plot:vanilla}
\end{figure}


\subsection{Offline Reinforcement Learning}


We train our agents on PyBullet Offline Reinforcement Learning datasets\footnote{ the generated PyBullet offline reinforcement learning examples are available from: \url{https://github.com/takuseno/d4rl-pybullet}}. These datasets are a frozen set of experiences generated from a SAC agent under different setups. The offline datasets used for each of the competing agents are identical. Furthermore the performance shown below is based on the agent interacting directly with the environment for evaluation purposes, and is not used for training. We compare the efficacy of our approach on three different sets of imperfect demonstration data, these being (1) completely random behavior policy, (2) a partially trained, medium scoring policy, and (3) a samples of experiences generated when constructed the partially trained agent. We trained behavior policies using Soft Actor-Critic algorithm \cite{sac}. In all cases, random data was generated by running a uniform at random policy in the environment. Mediocre data was generated by training SAC agent for 1 million timesteps for each of the environment to yield ``medium'' agents, at which case the agent policy was frozen, and the data was generated against this agent. Mixed data was generated by leveraging the experiences created in the training of the mediocre agent. We used the same datasets for evaluating different algorithms to maintain uniformity across results. 

We provide the average return and the standard deviation across all training runs in Table \ref{appendix:exp}.

\begin{figure*}
\centering
\begin{tabular}{c}
{\includegraphics[width=0.31\textwidth, height=6cm]{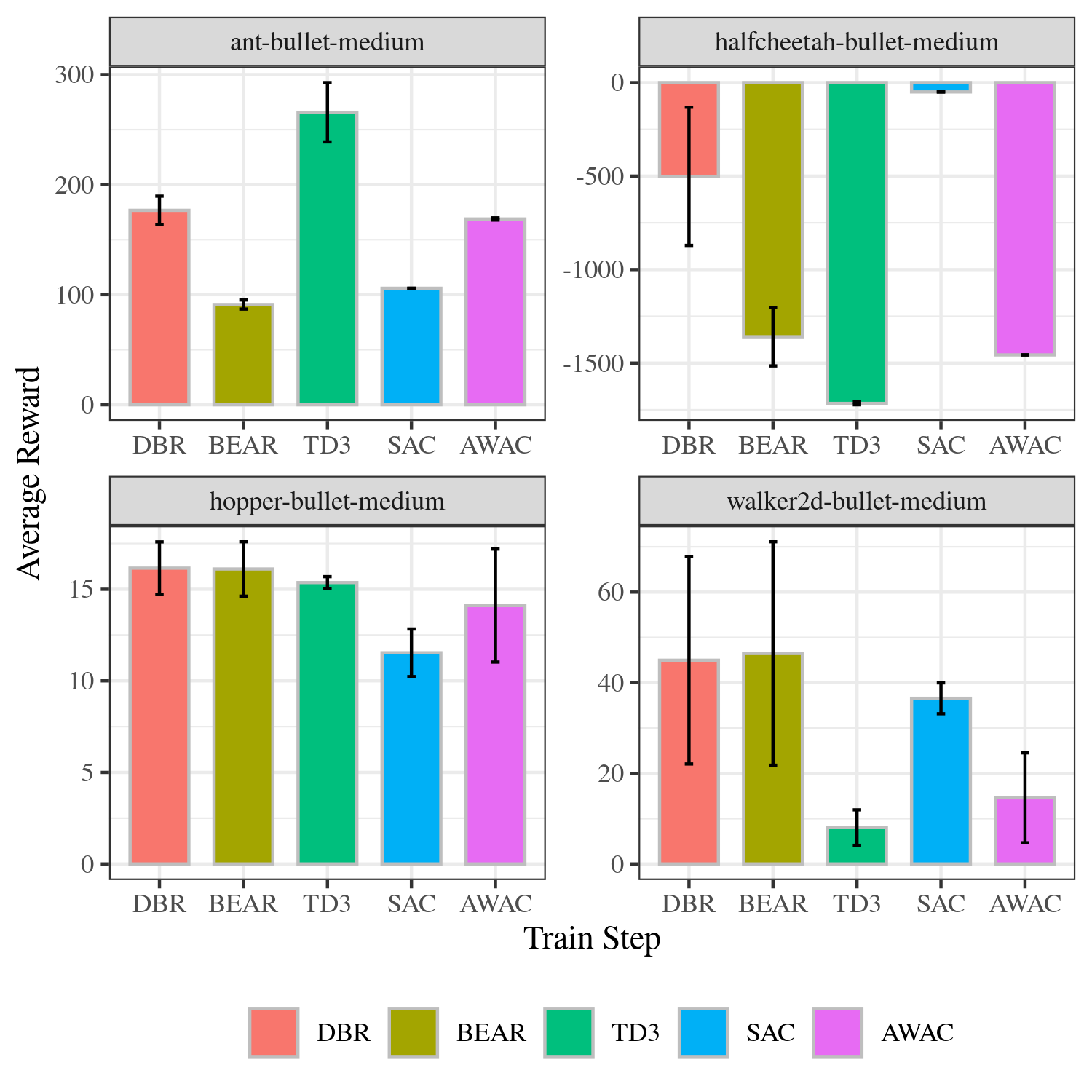} }
{\includegraphics[width=0.31\textwidth, height=6cm]{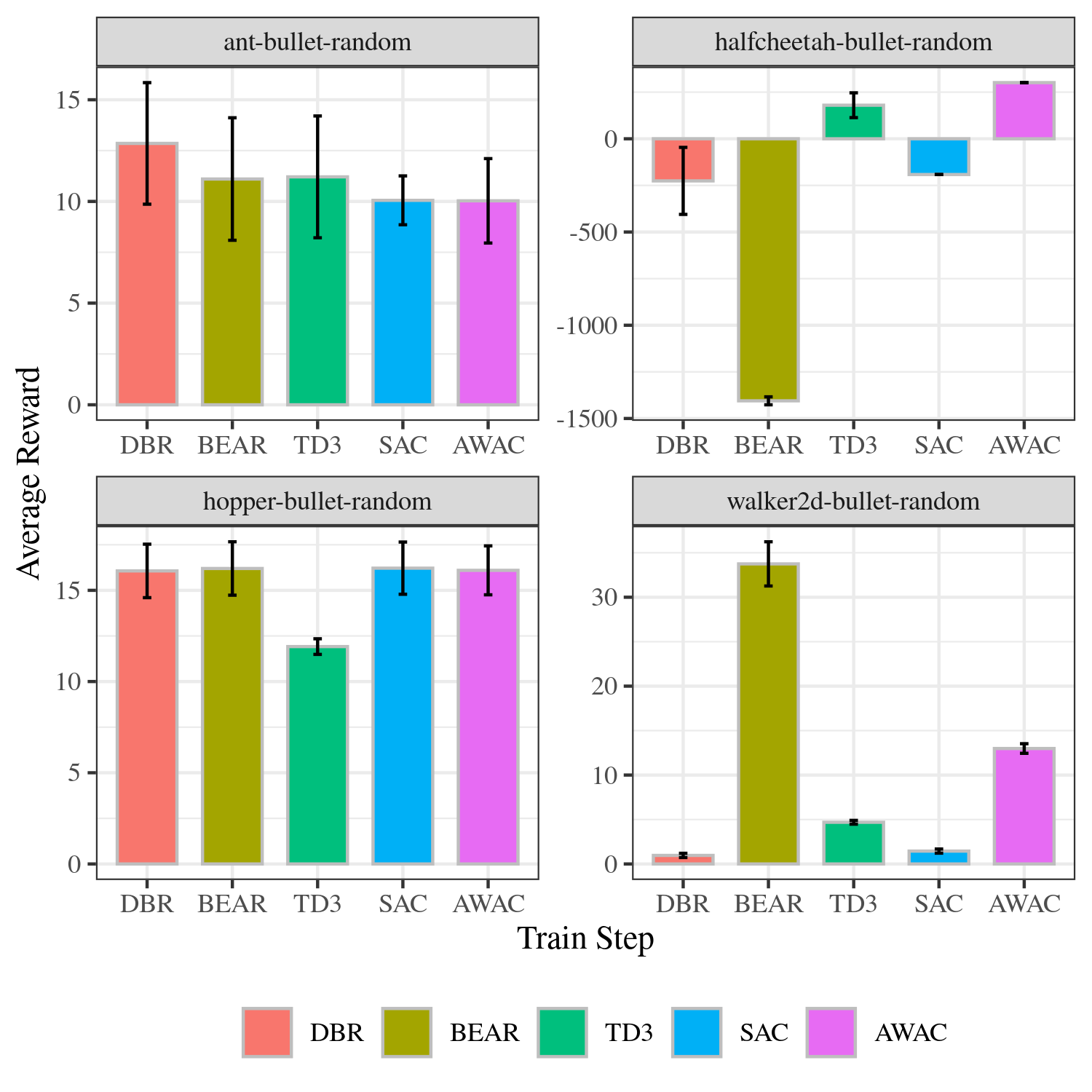} }
{\includegraphics[width=0.31\textwidth, height=6cm]{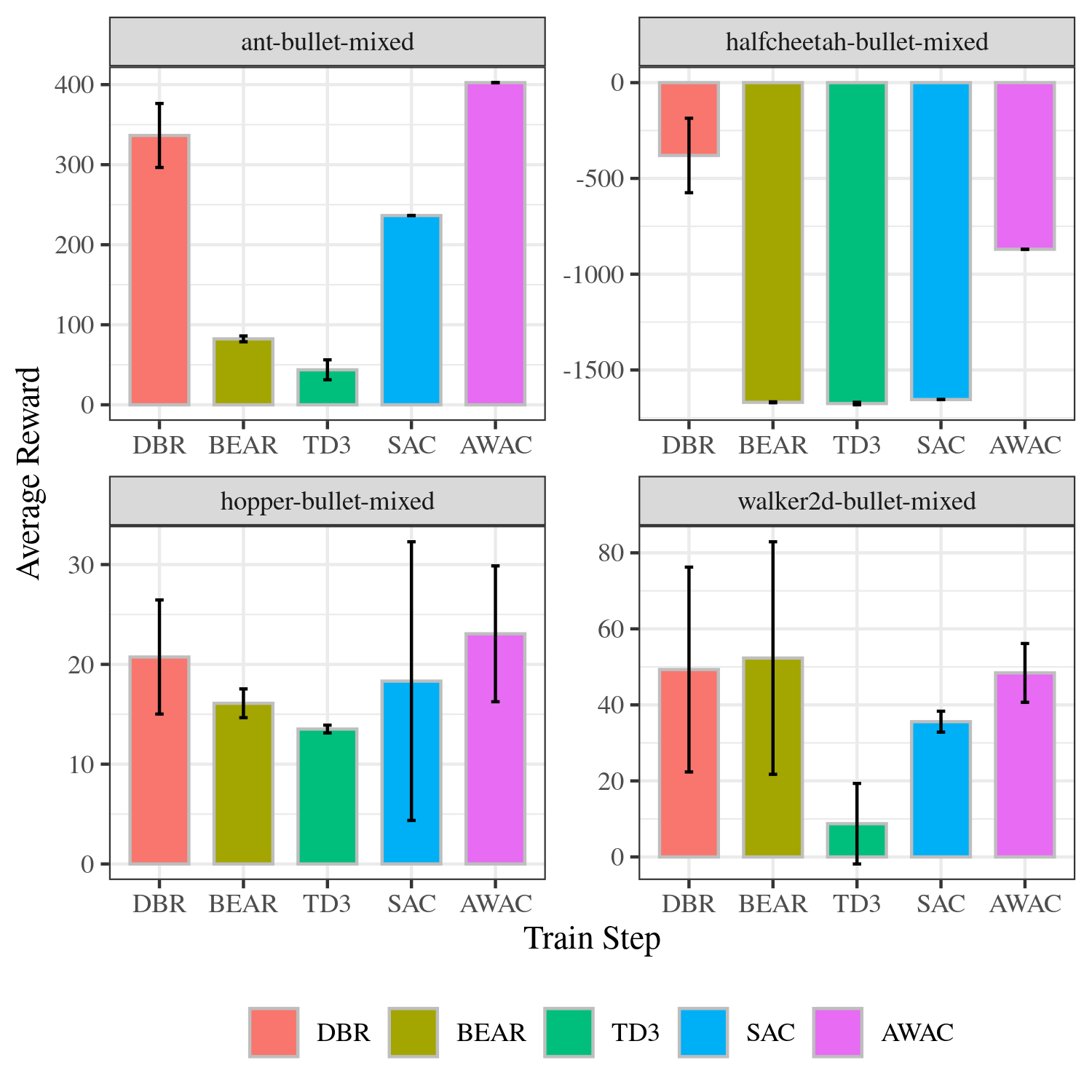} }
\end{tabular}
\caption{Comparison over PyBullet Control environments trained using offline datasets only, using (left) partially trained, medium scoring policy (center) randomly generated actions, (right) mixture of agent experiences gathered during training.}
\label{bar:offline}
\end{figure*}

\clearpage
\renewcommand{\arraystretch}{1.8}
\begin{landscape}
\begin{table}[p]
\caption{Results in all tasks for the different environments. The average return and the standard deviation during evaluation is presented. Discrete environments in Connect-4 for BEAR and DBR were implemented using Soft-Actor Critic discrete variations \cite{sac-discrete}}. 
\centering
\begin{tabular}{lcccccc}
\toprule
\textbf{Tasks/Algorithms}             & DBR                 & BEAR & TD3  & SAC & AWAC & DDQN \\ \midrule
Connect 4 – Random                    & $ 0.982 \pm 0.0158              $ & $0.967 \pm 0.116$ & $-            $ &$ -          $  &$ -          $   & $0.589\pm0.161 $  \\
Connect 4 – MCTS 10                   & $ 0.912\pm0.0793                $ & $0.75\pm0.187   $ & $-            $ &$ -          $  &$ -          $   & $0.214\pm0.161 $  \\ \hline
Ant-bullet (Offline – Medium)         & $ 175\pm2                       $ & $91\pm18.25     $ & $266\pm53.9     $ &$ 106\pm2      $  &$ 169\pm2      $   & $-           $  \\
Halfcheetah-bullet (Offline – Medium) & $ -1539 \pm 8.2$ & $ -1359\pm312   $ & $ -1716\pm14.7  $ &$  -49.9\pm9.2 $  &$  1456\pm6.4  $   & $-           $  \\
Hopper-bullet (Offline – Medium)      & $ 17.3\pm6                      $ & $16.1\pm2.97    $ & $15.4\pm0.65    $ &$ 11.5\pm2.6   $  &$ 14.1\pm6.17  $   & $-           $  \\
Walker2d-bullet (Offline – Medium)    & $ 33.3\pm4.8                    $ & $46.5\pm49.3    $ & $8.03\pm7.83    $ &$ 36.6\pm6.81  $  &$ 14.6\pm19.8  $   & $-           $  \\ \hline
Ant-bullet (Offline – Mixed)          & $ 292\pm0.255                   $ & $82.4\pm7.32    $ & $43.7\pm24.9    $ &$ 236 \pm 0.153$  &$ 403\pm0.119  $   & $-           $  \\
Halfcheetah-bullet (Offline – Mixed)  & $  -1694\pm9.3                  $ & $ -1669\pm14.78 $ & $ -1675 \pm 19.2$ &$  -1654\pm8.7 $  &$  -870\pm3.1  $   & $-           $  \\
Hopper-bullet (Offline – Mixed)       & $ 12.1\pm3.38                   $ & $16.1\pm2.88    $ & $13.5\pm0.783   $ &$ 18.3\pm27.9  $  &$ 23.1\pm13.6  $   & $-           $  \\
Walker2d-bullet (Offline – Mixed)     & $ 39.3\pm7.99                   $ & $52.3\pm61.2    $ & $8.74\pm21.2    $ &$ 35.6\pm5.51  $  &$ 48.4\pm15.5  $   & $-           $  \\ \hline
Ant-bullet (Offline – Random)         & $ 13.6\pm5.73                   $ & $11.1\pm6.02    $ & $11.2\pm5.99    $ &$ 10.1\pm2.4   $  &$ 10.0 \pm 4.15$   & $-           $  \\
Halfcheetah-bullet (Offline – Random) & $  -226 \pm 359                 $ & $ -1405\pm42.7  $ & $180\pm133      $ &$  -192 \pm 8.2$  &$ 301\pm4.6    $   & $-           $  \\
Hopper-bullet (Offline – Random)      & $ 16.6\pm2.88                   $ & $16.2\pm2.93    $ & $11.9\pm0.856   $ &$ 16.2\pm2.86  $  &$ 16.1\pm2.68  $   & $-           $  \\
Walker2d-bullet (Offline – Random)    & $ 0.822\pm0.490                 $ & $33.7\pm4.97    $ & $4.69\pm0.426   $ &$ 1.45\pm0.465 $  &$ 13.0\pm1.08  $   & $-           $  \\ \hline
Ant-pybullet-v0                       & $2136\pm6.5                     $& $567\pm9.3       $& 9$8.9\pm3.4      $& $2302\pm7.5    $ & $547\pm9.2     $  & -$            $ \\
Halfcheetah-pybullet-v0               & $ 919\pm9.4                     $ & $19.3\pm18      $ & $2172\pm4.5     $ &$ 1370\pm3.2   $  &$ 383\pm2.8    $   & $-           $  \\
Hopper-pybullet-v0                    & $ 2435\pm923                    $ & $33.4\pm1.83    $ & $2470\pm9.3     $ &$ 2031\pm7.4   $  &$ 794\pm6.1    $   & $-           $  \\
Walker2d-pybullet-v0                  & $ 1837\pm758                    $ & $47.0\pm16.9    $ & $1611\pm117     $ &$ 1721\pm7.39  $  &$ 6.75\pm5.82  $   & $-           $  \\ 
\bottomrule
\end{tabular}
\end{table}%
\end{landscape}
\renewcommand{\arraystretch}{1}

\clearpage

\subsubsection{Performance on Medium-Quality Data}

We first discuss the evaluation of condition with ``mediocre'' data, as this is resembles settings where training on offline data would be the most useful. One million transitions from a partially trained policy were collected, to simulate imperfect demonstration data or data from a mediocre prior policy. Compared with BEAR, it consistently performed similarly or noticeably stronger as shown in Figure \ref{bar:offline}(a). This scenario is the most relevant, as random data may not have adequate exploration to learn a good policy. We observe in the two scenarios where SAC or TD3 outperformed DBR, DBR was the clear second-best algorithm in the ``ant-bullet'' and ``halfcheetah-bullet'' environments respectively.

\subsubsection{Performance on Random and Mixed-Quality Data}

In Figure \ref{bar:offline}, we show the performance of each method when trained on data from a random policy (b) and a mixed policy (c). The ``random'' dataset was created from generating data made from an agent randomly sampling from the environment's action space, whereas ``mixed'' data was creating from sampling the experiences of an agent during training. In both cases, our method DBR, achieves good results, generally performing comparable or better than BEAR and other approaches, with the exception of ``walker2d-bullet-random'' and ``hopper-bullet-mixed'' environments. These results are generally consistent with the expectation that BEAR, DBR would be more robust to the dataset composition. 


\subsection{Analysis of DBR}

\label{ref:abl}

\begin{figure}
\centering
\begin{tabular}{c}
{\includegraphics[width=0.4\textwidth]{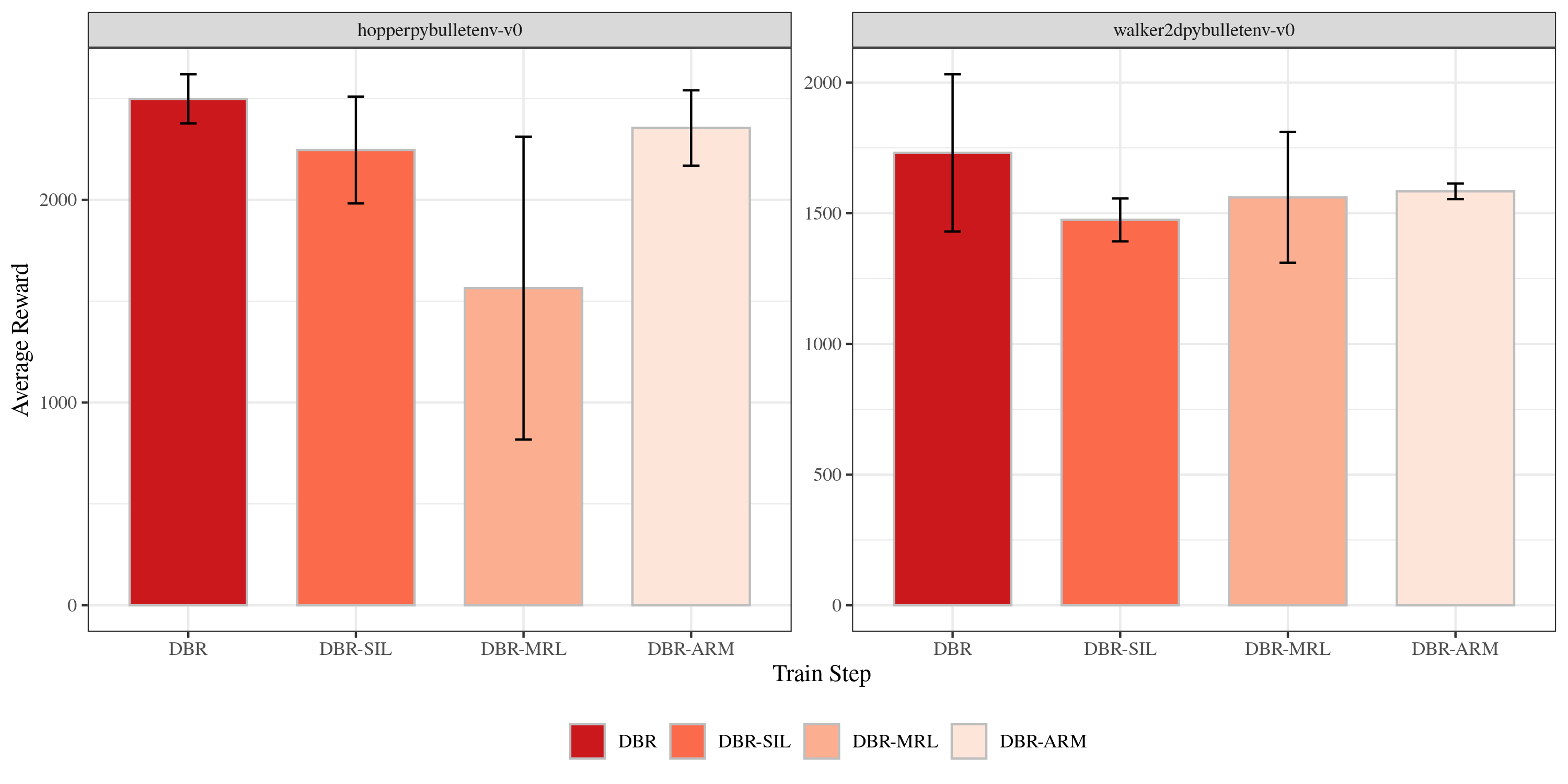}}
{\includegraphics[width=0.4\textwidth]{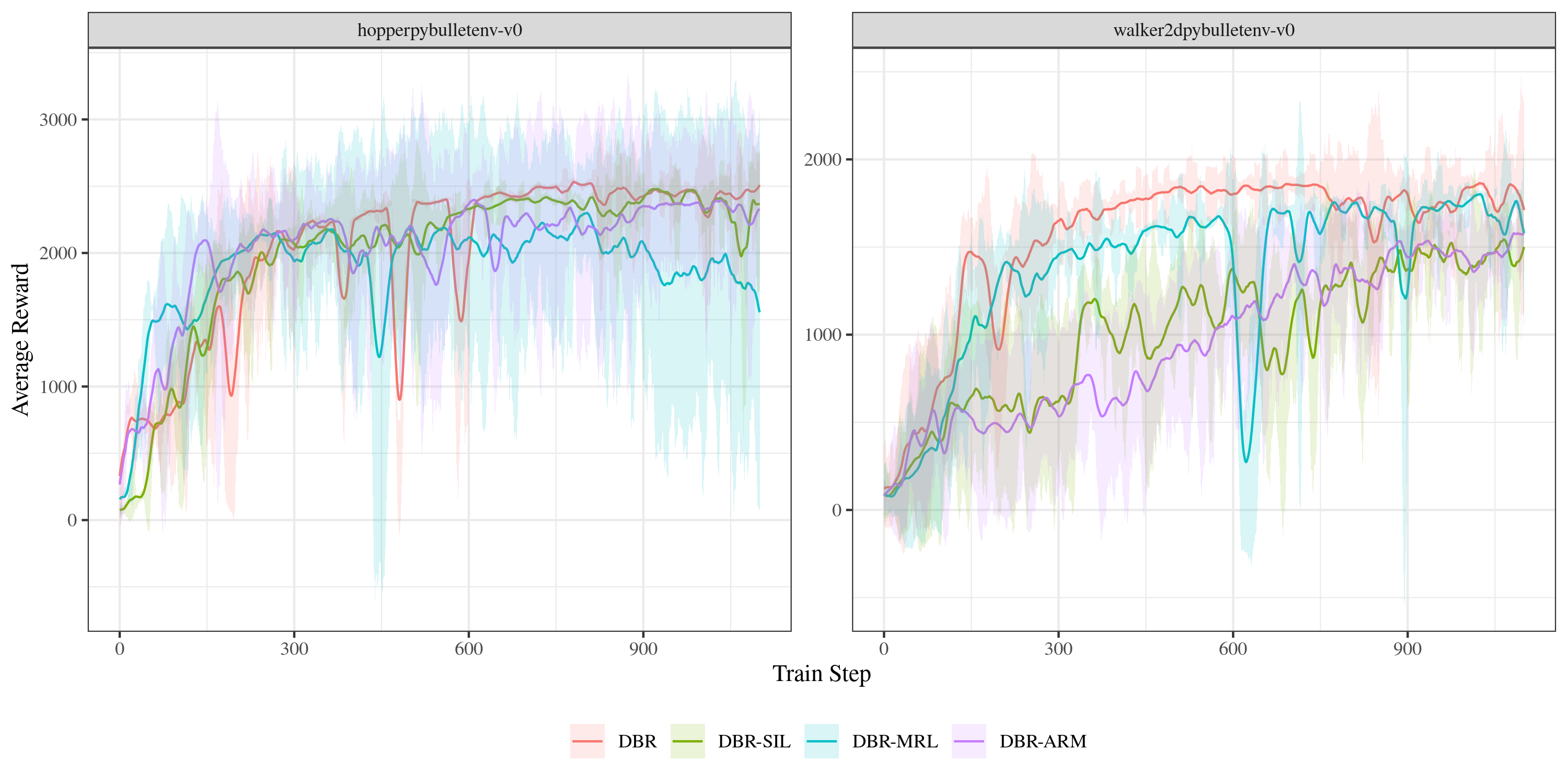}}
\end{tabular}
\caption{Comparing the performance over different approaches. (left) We report the mean over the last 10 evaluation steps, where each evaluation point is average over 1000 episodes. Error bars represent the standard deviation over these rollouts (right) demonstrates the performance across the training iteration.}
\label{plot:abl}
\end{figure}

We examine the effect of using Advantage-Based Regret Minimization and Self-imitation learning variation of clipped advantage with a single behavior policy and the Munchhausen Reinforcement Learning approach using an augmented reward based on the behavior policy, with an entropy regularized approach as described in Section \ref{method:dbr} to examine the effects of implicit regularization. As shown in Figure \ref{plot:abl}, we can empirically observe the stability of our approach of dual behavior policies over a range of different modifications, whereby our choice to use the ARM variation of clipped advantage is demonstrated through better performance of a range of environments. 

We observe the degenerative nature of MRL, which suggests that for this particular application, it is worthwhile explicitly regularizing the behavior policy rather than implicitly, especially when compared with both the SIL and ARM approaches. We also justify our approach for using the clipped advantage variant used in ARM over SIL empirically in these experiments, as the ablation approach using ARM outperforms SIL.

\section{Discussion and Future Work}

We have discussed the formulation of an offline reinforcement learning approach which relaxes the constraint on its behavior policy in a dynamic manner. The key insight is that DBR removes the fixed constraint, and instead allows for a dynamic way to control the explore-exploit trade-off through consideration of the positive and negative clipped behavior policies. This appears to allow our approach to generalize and perform experimentally over a variety of dataset compositions, learning on random, medium-quality and expert data. Furthermore we have justified our choice of clipped advantage both theoretically by demonstrating the relationship with regret minimization and empirically by comparing it with the variation used in self-imitation learning. In conclusion we have outlined an extension to offline reinforcement learning with a dynamic constraint to control for exploration. One direction of future work would be to extend this approach to large scaling problems involving multi-agents which need to cooperate or are adversarial. Of interest is the construction of different behaviors to capture the different nuances in how agents may interact with each other and the environment. If capturing this nuance is possible; how might behaviors be transferred from one context to another efficiently? Is it possible to learn and constrain agents to multiple behaviors at once which may be generated in a heuristic manner? In this way, reinforcement learning can effectively learn different behaviors from large-scale off-policy datasets, enabling rapid progress in terms of accuracy and generalization seen in recent years within the supervised learning fields. 


\bibliography{acml20}

\begin{thebibliography}{10}

\bibitem{abdolmaleki2018maximum}
Abbas Abdolmaleki, Jost~Tobias Springenberg, Yuval Tassa, Remi Munos, Nicolas
  Heess, and Martin Riedmiller.
\newblock Maximum a posteriori policy optimisation.
\newblock In {\em International Conference on Learning Representations}, 2018.

\bibitem{Bennett07thenetflix}
James Bennett, Stan Lanning, and Netflix Netflix.
\newblock The netflix prize.
\newblock In {\em In KDD Cup and Workshop in conjunction with KDD}, 2007.

\bibitem{mcts}
Guillaume Chaslot, Sander Bakkes, Istvan Szita, and Pieter Spronck.
\newblock Monte-carlo tree search: A new framework for game ai.
\newblock In {\em Proceedings of the Fourth AAAI Conference on Artificial
  Intelligence and Interactive Digital Entertainment}, AIIDE'08, page
  216–217. AAAI Press, 2008.

\bibitem{sac-discrete}
Petros Christodoulou.
\newblock Soft actor-critic for discrete action settings.
\newblock {\em arXiv preprint arXiv:1910.07207}, 2019.

\bibitem{d4rl}
Justin Fu, Aviral Kumar, Ofir Nachum, George Tucker, and Sergey Levine.
\newblock D4rl: Datasets for deep data-driven reinforcement learning, 2020.

\bibitem{fudiag}
Justin Fu, Aviral Kumar, Matthew Soh, and Sergey Levine.
\newblock Diagnosing bottlenecks in deep q-learning algorithms.
\newblock In Kamalika Chaudhuri and Ruslan Salakhutdinov, editors, {\em
  Proceedings of the 36th International Conference on Machine Learning},
  volume~97 of {\em Proceedings of Machine Learning Research}, pages
  2021--2030. PMLR, 09--15 Jun 2019.

\bibitem{modeseeking}
Seyed Kamyar~Seyed Ghasemipour, Richard Zemel, and Shixiang Gu.
\newblock A divergence minimization perspective on imitation learning methods.
\newblock In {\em Conference on Robot Learning}, pages 1259--1277. PMLR, 2020.

\bibitem{sac}
Tuomas Haarnoja, Aurick Zhou, Pieter Abbeel, and Sergey Levine.
\newblock Soft actor-critic: Off-policy maximum entropy deep reinforcement
  learning with a stochastic actor.
\newblock In Jennifer Dy and Andreas Krause, editors, {\em Proceedings of the
  35th International Conference on Machine Learning}, volume~80 of {\em
  Proceedings of Machine Learning Research}, pages 1861--1870,
  Stockholmsmässan, Stockholm Sweden, 10--15 Jul 2018. PMLR.

\bibitem{ddqn}
Hado~van Hasselt, Arthur Guez, and David Silver.
\newblock Deep reinforcement learning with double q-learning.
\newblock In {\em Proceedings of the Thirtieth AAAI Conference on Artificial
  Intelligence}, AAAI'16, page 2094–2100. AAAI Press, 2016.

\bibitem{arm}
Peter Jin, Kurt Keutzer, and Sergey Levine.
\newblock Regret minimization for partially observable deep reinforcement
  learning.
\newblock In {\em International conference on machine learning}, pages
  2342--2351. PMLR, 2018.

\bibitem{bear}
Aviral Kumar, Justin Fu, Matthew Soh, George Tucker, and Sergey Levine.
\newblock Stabilizing off-policy q-learning via bootstrapping error reduction.
\newblock In {\em Advances in Neural Information Processing Systems}, pages
  11784--11794, 2019.

\bibitem{litoward}
Lihong Li, Remi Munos, and Csaba Szepesvari.
\newblock {Toward Minimax Off-policy Value Estimation}.
\newblock In Guy Lebanon and S.~V.~N. Vishwanathan, editors, {\em Proceedings
  of the Eighteenth International Conference on Artificial Intelligence and
  Statistics}, volume~38 of {\em Proceedings of Machine Learning Research},
  pages 608--616, San Diego, California, USA, 09--12 May 2015. PMLR.

\bibitem{lillicrap2015continuous}
Timothy~P Lillicrap, Jonathan~J Hunt, Alexander Pritzel, Nicolas Heess, Tom
  Erez, Yuval Tassa, David Silver, and Daan Wierstra.
\newblock Continuous control with deep reinforcement learning.
\newblock In {\em International Conference on Learning Representations}, 2016.

\bibitem{mnih2013playing}
Volodymyr Mnih, Koray Kavukcuoglu, David Silver, Alex Graves, Ioannis
  Antonoglou, Daan Wierstra, and Martin Riedmiller.
\newblock Playing atari with deep reinforcement learning.
\newblock {\em arXiv preprint arXiv:1312.5602}, 2013.

\bibitem{murphy2001marginal}
Susan~A Murphy, Mark~J van~der Laan, James~M Robins, and Conduct Problems
  Prevention~Research Group.
\newblock Marginal mean models for dynamic regimes.
\newblock {\em Journal of the American Statistical Association},
  96(456):1410--1423, 2001.

\bibitem{awac}
Ashvin Nair, Murtaza Dalal, Abhishek Gupta, and Sergey Levine.
\newblock Accelerating online reinforcement learning with offline datasets,
  2020.

\bibitem{vaepolsearch}
Gerhard Neumann.
\newblock Variational inference for policy search in changing situations.
\newblock In {\em Proceedings of the 28th International Conference on
  International Conference on Machine Learning}, ICML'11, page 817–824,
  Madison, WI, USA, 2011. Omnipress.

\bibitem{sil}
Junhyuk Oh, Yijie Guo, Satinder Singh, and Honglak Lee.
\newblock Self-imitation learning.
\newblock In Jennifer Dy and Andreas Krause, editors, {\em Proceedings of the
  35th International Conference on Machine Learning}, volume~80 of {\em
  Proceedings of Machine Learning Research}, pages 3878--3887,
  Stockholmsmässan, Stockholm Sweden, 10--15 Jul 2018. PMLR.

\bibitem{mrl}
Nino Vieillard, Olivier Pietquin, and Matthieu Geist.
\newblock Munchausen reinforcement learning.
\newblock {\em Advances in Neural Information Processing Systems}, 33, 2020.

\bibitem{brac}
Yifan Wu, George Tucker, and Ofir Nachum.
\newblock Behavior regularized offline reinforcement learning.
\newblock {\em arXiv preprint arXiv:1911.11361}, 2019.

\bibitem{yu2020bdd100k}
Fisher Yu, Haofeng Chen, Xin Wang, Wenqi Xian, Yingying Chen, Fangchen Liu,
  Vashisht Madhavan, and Trevor Darrell.
\newblock Bdd100k: A diverse driving dataset for heterogeneous multitask
  learning.
\newblock In {\em Proceedings of the IEEE/CVF Conference on Computer Vision and
  Pattern Recognition}, pages 2636--2645, 2020.

\end{thebibliography}
\bibliographystyle{plain}

\end{document}